\documentclass[reqno,9pt]{amsart}
\usepackage{hyperref}
\usepackage{graphicx}
\usepackage{slashed}
\usepackage{amscd}
\usepackage{tensor}
\usepackage{amssymb}
\usepackage{esint}
\usepackage{enumitem}
\usepackage{accents}
\usepackage[dvipsnames]{xcolor}
\usepackage[utf8]{inputenc}
\usepackage[off]{auto-pst-pdf}
\usepackage{pst-grad}
\usepackage{pst-plot}
\usepackage[mathscr]{eucal}
\usepackage{tagging}
\droptag{NotReady}

\numberwithin{equation}{section}
\allowdisplaybreaks[1]

\newtheorem{Def}{Definition}
\newtheorem{Thm}[Def]{Theorem}

\newtheorem{Lemma}[Def]{Lemma}

\newtheorem{Example}[Def]{Example}

\newcommand{\beq}{\begin{equation}}
\newcommand{\eeq}{\end{equation}}
\newcommand{\Proof}{\begin{proof}}
\newcommand{\QED}{\end{proof} \noindent}
\newcommand{\QEDrem}{\ \hfill $\Diamond$}

\DeclareFontFamily{OT1}{rsfso}{}
\DeclareFontShape{OT1}{rsfso}{m}{n}{ <-7> rsfso5 <7-10> rsfso7 <10-> rsfso10}{}
\DeclareMathAlphabet{\mycal}{OT1}{rsfso}{m}{n}

\newcommand{\iD}{\raisebox{0.125em}{\tiny $\blacktriangleright$}}

\newcommand{\bei}{\begin{itemize}[label=$\circ$,itemsep=.5em,leftmargin=*]}
\newcommand{\beii}{\begin{itemize}[label=$\rightarrow$,itemsep=1.5em,topsep=.5em,leftmargin=*]}
\newcommand{\eni}{\end{itemize}}
\newcommand{\enii}{\end{itemize}}

\definecolor{darkblue}{RGB}{0,91,163}

\newcommand{\cbb}[1]{   
\begin{itemize}[label=\iD,topsep=.5em,itemsep=.5em,leftmargin=*]
\item \textsf{\small #1}
    \begin{itemize}[label=-,itemsep=.2em,topsep=.2em,leftmargin=*]
}
\newcommand{\cbe}{        
    \end{itemize}
\end{itemize}
}

\usepackage{multicol}

\newcommand{\Tcomp}{{T_{\textrm{comp}}}}

\newcommand{\Phys}{\Upsilon}

\usepackage{lipsum}

\begin{document}

\begin{center}
\Huge

If consciousness is dynamically relevant,\\[.1em]
artificial intelligence isn't conscious 

\end{center}

\vspace{.7cm}

\begin{center}
\small Johannes Kleiner$^{1,2,3,4}$ and Tim Ludwig$^{5}$
\end{center}
\vspace*{.3cm}
\centerline{$^1$Munich Center for Mathematical Philosophy, LMU Munich}
\centerline{$^2$Munich Graduate School of Systemic Neurosciences, LMU Munich}
\centerline{$^3$Institute for Psychology, University of Bamberg}
\centerline{$^4$Association for Mathematical Consciousness Science}
\centerline{$^5$Institute for Theoretical Physics, Utrecht University}\vspace{-.3em}
\centerline{\small Princetonplein 5, 3584 CC Utrecht, The Netherlands}

\vspace{.7cm}

\begin{quote}
    \textsc{Abstract.} 
    We demonstrate that if consciousness is relevant for the temporal evolution of a system's states---that is, if it is \emph{dynamically relevant}---then 
    AI systems cannot be conscious. That is because AI systems run on CPUs, GPUs, TPUs or other processors which have been designed and verified to adhere to computational dynamics that systematically preclude or suppress deviations. The design and verification preclude or suppress, in particular, potential consciousness-related dynamical effects, so that if consciousness is dynamically relevant, AI systems cannot be conscious.
\end{quote}


\vspace{.7cm}

\begin{multicols}{2}
The question of whether artificial intelligence (AI) systems are conscious has emerged as one of critical scientific, philosophical, and societal concern. 
While empirical support to differentiate theories of consciousness is still nascent and while current measures of consciousness (the simplest example of which is interpretation of verbal reports) cannot justifiably be applied to AI systems, our best hope for reliable answers is to link AI's potential for consciousness with fundamental properties of conscious experience that have empirical import or philosophical credibility.

Significant progress in this regard has already been achieved. In~\cite{chalmers2023could}, David Chalmers assesses evidence for or against AI consciousness based on an extensive array of features that a system or organism might possess or lack, such as self-report, conversational ability, general intelligence, embodiment, world or self-models, recurrent processing, or the presence of a global workspace.
In~\cite{wiesecould}, Wanja Wiese proposes a criterion for distinguishing between conscious and non-conscious AI, anchored in the desiderata of the neuroscientific Free Energy Principle.%
    \footnote{
    These are examples of research whose aim is to evaluate \emph{whether} AI systems of the more recent form are or can be conscious. 
    Other interactions between AI research and consciousness science include the use of AI inspired tools and concepts to model consciousness, for example~\cite{blumstheory,ji2023sources,juliani2022perceiver}, and studies of how models of consciousness might help to build better AI, for example~\cite{blum2023theoretical,juliani2022link,mollo2023vector}.
    The question of whether machines in general can be conscious has guided much of the debate in philosophy
    of mind over the previous decades, cf.~\cite{block1980troubles,bronfman2021will,chalmers2016singularity,CLANCEY1993313,clark1998being,dennett1991consciousness,haugeland1989artificial,holland2003machine,penrose1991emperor,searle1980minds,SMOLIAR1991295,tegmark2018life,alanturing}, among others.
    }

In this paper, we propose a result of similar nature, which however does not rely on system features and how they relate to consciousness, but on a general property of consciousness: \emph{dynamical relevance}. Here, \emph{dynamical} refers to the temporal evolution (the dynamics) of a system's physical states. Consciousness is \emph{relevant} to a system's time evolution if the time evolution with consciousness differs from the time evolution without consciousness. Whether or not consciousness is dynamically relevant depends on the theory of consciousness under consideration, and in how far this theory implements consciousness-dependent changes of the dynamical evolution, as compared to a physical theory that addresses the same states.

What sets AI systems apart in the context of consciousness is not the specific computational architecture that is employed; architectures that closely resemble the mammalian brain's computational structure can arguably also be used, after all~\cite{friston2022designing}. Instead, the distinctive aspect is the hardware on which an AI architecture operates, namely CPUs, GPUs, TPUs, or other processors. This hardware is designed and verified to ensure that the physical dynamics evolve precisely as described by a computational theory during what is known as \emph{functional} and \emph{post-silicon verification}. These verification processes ensure that the physical design of the chip (the layout of integrated circuits in terms of semiconductors), as well as the actual physical product (the processing unit after production), yield dynamics exactly as specified by the computational theory. Any dynamical effects that violate the specification of this theory are excluded or dynamically suppressed by error correction.

The \emph{intuition} behind our result is summarised below. The objective of the paper is to delineate all concepts involved in this intuition carefully, so as to present a theorem 
that underwrites the intuition both in scope and precision.
\begin{enumerate}[label=(A\arabic*),leftmargin=2.7em]
    \item Verification of processing units ensures that any dynamical effects that change the computational dynamics of a processing unit are precluded or suppressed.\label{intuition:very}
    \item If consciousness is dynamically relevant, and AI systems are conscious, then there are dynamical effects that change the computational dynamics of an AI system.\label{intuition:consc}
    \item AI systems run on processing units.\label{intuition:run}
    \rule[.2em]{10em}{.1pt}
    \item[(C)] If consciousness is dynamically relevant, AI systems cannot be conscious.\label{intuition:conclusion}
\end{enumerate}
The conclusion~(C) follows because qua~\ref{intuition:run} and~\ref{intuition:very}, verification ensures that any dynamical effects that change the computational dynamics of an AI system are precluded or suppressed. \ref{intuition:consc} states that if consciousness is dynamically relevant, 
and AI systems are conscious, then there are dynamical effects that change the computational dynamics of an AI system. Therefore, if consciousness is dynamically relevant, then AI systems cannot be conscious.
The crucial work of the formalisation we introduce below is to make sure this reasoning is also sound if consciousness' dynamical effects apply on a ``level below'' the computational level.

In a nutshell, this paper shows that if consciousness makes a difference to how a system evolves in time---as it should if consciousness is to have any evolutionary advantage, for example---then any system design which systematically precludes or suppresses diverging dynamical effects systematically precludes or suppresses the system from being conscious.

Before embarking on the formal research that puts the above reasoning on solid ground, we focus on the new concept of dynamical relevance: we explain it in more detail in Section~\ref{sec:what_is}; and, in Section~\ref{sec:why}, we argue why it should be assumed to hold for consciousness.

\section{What is Dynamical Relevance?}\label{sec:what_is}

In this section, we explain and illustrate the concept of \emph{dynamical relevance}, and discuss its relation to other concepts in the context consciousness.

Dynamical relevance is a relational concept. It describes how something, for example a property, relates to the dynamics of a system, as described by a theory. If that ``something'' is relevant for the dynamics of the system, then we call it \emph{dynamically relevant}. In contrast, if that ``something'' is not relevant for the dynamics of the system, then we call it \emph{not dynamically relevant} or \emph{dynamically irrelevant}. Before applying dynamical relevance to consciousness, let us give two examples to further illustrate the concept.

\subsection*{Example 1: A moving car} As an intuitive first example, we consider a hypothetical theory for a moving car.%
    \footnote{We thank Wanja Wiese for suggesting this example when discussing our manuscript.}
The theory predicts, we presume, how the car behaves as forces are applied to it. In particular, it describes which dynamical trajectory the car takes on a parking lot as forces are applied to its steering wheel and its brake and gas pedals for a given initial position and velocity.
    
How much load we add to the car is not predicted by the moving-car theory. If one puts a heavy box into the trunk of the car, however, the car's dynamical trajectory will be different from its dynamical trajectory with an empty trunk. This difference might be small and hard to notice or large and easy to notice; for example, in the case of a Moose test, a heavy box in the trunk could make the difference between tipping over and not tipping over. In any case, as the load of the car makes a difference to the dynamics of the car, we conclude that the car's load is dynamically relevant for the car, as described by the moving-car theory.

The colour of the car's seats is also not predicted by the moving-car theory. For example, they could be coloured in black, blue, or red. In contrast to the car's load, however, the car's dynamical trajectory will be the same for all seat colours. Thus, as the seat colour doesn't make a difference to the dynamics of the car, we conclude that the seat colour is dynamically irrelevant for the car, as described by the moving-car theory.

To summarise, for the hypothetical moving-car theory outlined above, the car's load is dynamically relevant, whereas the seats' colour is dynamically irrelevant. We emphasise that the specification of the theory is important. For another, more elaborate, moving-car theory that takes into account the driver and their psychology for the prediction of the car's dynamical trajectory, the seats' colour might very well make a difference for the dynamics of the car and, thus, be dynamically relevant.

\subsection*{Example 2: An electrical circuit}
As a more scientific example, we consider an electrical circuit. In an electrical circuit, voltages and charge currents are typically described by electrical circuit theory. For example, Ohm's law $V = R \cdot I$ relates the voltage drop $V$ across an electrical resistor with resistance $R$ to the charge current flow $I$ through the resistor. Besides the resistor, the electrical capacitor is another important circuit element. A capacitor stores electrical charge $Q$, when a voltage $V$ is applied to it; the capacitor's capacity $C$ determines the amount of charge that is stored for a given voltage $Q=CV$.

Based on the two circuit elements, resistor and capacitor, one can build a simple electrical circuit: a so called $RC$-circuit, where a capacitor is effectively connected to itself but only via the resistor. When the capacitor is initially charged up to the voltage $V_0$, it will decay on a timescale $\tau= RC$; explicitly, $V(t) = V_0\, e^{-t/\tau}$. So, the resistance $R$ and the capacity $C$ are both relevant for the dynamics of the voltage $V$. Thus, resistance and capacity are dynamically relevant for the voltage dynamics in an $RC$-circuit, as described by circuit theory. Note that this is an example for dynamical relevance within a theory; voltage, resistance, and capacity are all concepts within circuit theory.

We can also apply the concept of dynamical relevance beyond circuit theory. The resistance of a resistor $R$ depends on the temperature of the resistor $T$. Note that temperature is a concept from thermodynamics but not from circuit theory. Nevertheless, we can say that the temperature is relevant for the resistance. In turn, the temperature is dynamically relevant for the voltage in an $RC$-circuit. In contrast, similar to the previous example, the resistor's colour coating is not relevant for its resistance. In turn, the resistor's colour coating is dynamically irrelevant for the voltage in an $RC$-circuit.

\subsection*{Dynamical relevance of consciousness}
Having clarified the concept of dynamical relevance, we can now discuss its application in consciousness science.

Dynamical relevance is used to describe the relation between a theory of consciousness and an underlying physical theory. In a nutshell, a theory of consciousness posits consciousness as dynamically relevant, if being conscious makes a difference for the time evolution of a system, as described by the physical theory. It is crucial to note that, similar to resistance and temperature in the circuit-theory example, the concept of dynamical relevance is meaningful whether or not consciousness is physical itself. In particular, it makes sense also if the theory of consciousness is a physicalist theory. Dynamical relevance is a condition on what consciousness does, not about what consciousness is.

A simple example of a theory of consciousness that posits consciousness to be dynamically relevant is a theory which proposes that consciousness is a specific physical cognitive function that would be absent if systems did not possess consciousness. Another simple example is a theory of consciousness which posits that consciousness is something non-physical and endows consciousness with a causal effect on physical states.

\subsection*{Relation to other Properties}

We have already indicated that consciousness can be dynamically relevant in both physicalist and non-physicalist ontologies. That is, it is \emph{ontologically neutral}. By endorsing dynamical relevance one is not committed to any specific ontology. 
As we will now show, dynamical relevance is furthermore implied by other (important) concepts in both physicalist and non-physicalist concepts. Therefore, dynamical relevance is a \emph{weaker} assumption than those concepts. It is easier to accept and less demanding than these other concepts.

In \emph{non-physicalist contexts}, dynamical relevance (of consciousness) is implied by a violation of an ontological assumption known as `causal closure of the physical' or `completeness of the physical'~\cite{sepmentalcausation}. This assumption states that for every physical effect, there are sufficient physical causes.

Dynamical relevance is implied by a violation of the causal closure of the physical, because if the physical is not causally closed in virtue of consciousness, there are physical effects at least one of whose jointly sufficient causes is consciousness---usually conceived of as a property or substance separate from the physical properties or substances in this context. But a cause makes a difference to the time-evolution of its effect. Hence it follows that consciousness makes a difference to the time evolution of some physical effects: the time evolution with consciousness differs from what it would have been without consciousness. Thus, if the physical is not causally closed in virtue of consciousness, consciousness is dynamically relevant.

In \emph{physicalist contexts}, dynamical relevance is implied by at least three concepts. First, it is implied by strong emergence. That is the case, because the ``fundamental higher-level causal powers''~\cite[Sect.\,4]{seppropertiesemergent}, which exist in the case of strong emergence, make a difference to the time evolution of the substrate states.

Second, dynamical relevance can also be implied by some forms of weak emergence. It is arguably implied, for example, by the information decomposition approach to causal emergence~\cite{mediano2022greater}. In this approach, even weak emergence induces downward causation. If downward causation implies that there are causal effects of the higher-level property on the lower-level property, then the higher-level property is dynamically relevant to the lower-level property.

Finally, dynamical relevance is also implied by the assumption that consciousness has intrinsic or functional value~\cite{cleeremans2022consciousness}, which motivates agents and guides their behaviour. That is the case because an agent's behaviour is part of the agent's dynamical trajectory. Therefore, if ``it is only in virtue of the fact that conscious agents `experience' things and `care' about those experiences that they are `motivated' to act in certain ways''~\cite[p.\,1]{cleeremans2022consciousness}, then consciousness is dynamically relevant.

We conclude this section with a pointer to the places in the manuscript where the precise definition of dynamical relevance is given. This is in Section~\ref{sec:tocs}, Definitions~\ref{def:CDRtheory} and~\ref{def:CDR}. Definition~\ref{def:CDRtheory} is epistemic. It defines the concept of dynamical relevance with respect to a theory of consciousness, relative to some underlying physical theory, independently of whether either of the theories is true.  Definition~\ref{def:CDR} then builds on this epistemic definition to provide an ontic definition. That is to say, this definition is about whether consciousness is actually dynamically relevant. What is crucial in Definition~\ref{def:CDR} is that it suffices that there is \emph{some} physical theory with respect to which the true theory of consciousness satisfies Definition~\ref{def:CDRtheory}. This is sufficient to prove our result, Theorem~\ref{thm}. Referencing the actual world is important in the context of this result because post-silicon verification is about what actually happens, once a processing unit has been manufactured, as well.

\section{Why Dynamical Relevance Should be Assumed to Hold}\label{sec:why}
While our result is predicated on dynamical relevance; it only applies \emph{if} dynamical relevance holds true. As we explain in this section, there is a very good, if not compelling, reason for assuming that dynamical relevance indeed does hold true for consciousness.

Consider, as a simple example, an experiment which relies on a subject's reports on her conscious experiences. Let us assume that the subject is shown some stimulus followed by a mask, and that she has to press a button to indicate whether she has consciously perceived the stimulus, or not, across various trials.
Throughout the trials, we might measure her EEG signal, so as to carry out an analysis that distinguishes EEG activity in the case of conscious perception from EEG activity in the case of unconscious perception. This analysis might target a theory of consciousness, so as to confirm or refute whether the difference in EEG signal is aliened with the theory's predictions or retrodictions about this case.

A necessary condition for such an analysis to be possible is that the report---the pressing of a button, in this case---\emph{can} depend on whether the subject has consciously perceived a stimulus, or not. Put in terms of the theory of consciousness that a study aims at, we may say: A necessary condition, for the above analysis to be possible, is that the report (or EEG data for that matter) depends on whether the subject is experiencing the stimulus consciously or not (according to the theory, if it were true). If the time-series of reports and EEG data does not depend on consciousness, the experiment cannot have any weight in supporting the theory.
In other words, the theory must posit consciousness as relevant to the report or EEG data (or both). And because report and EEG data are part of the dynamics of the physical, the theory must posit consciousness to be dynamically relevant. Dynamical relevance is a precondition for the experiment and the analysis to work as intended.

More generally, we may say that any empirical investigation of consciousness relies on \emph{measures of consciousness}~\cite{irvine2013measures} to infer the state of consciousness of a subject (some information about the subject's conscious experience, that is). An experiment may use objective measures of consciousness that rely on behavioural or neural markers, or subjective measures of consciousness that rely on a subject's reports about their conscious experience. Both types of measures rely on data that is part of the dynamics of the physical. And for a measure of consciousness to work as expected---to allow us to infer something about the state of consciousness of a subject---, consciousness must make a difference to the data that feeds into the measure. It must make a difference to the dynamics of the physical, and hence be dynamically relevant.

The same argument can be made not only for scientific investigations, but for any kind of intersubjective exploration of conscious experiences. Debating consciousness relies on certain dynamics of the vocal cord (among many other things), making art about consciousness makes use of behaviour. All of these cases are part of the dynamics of an organism, and if the dynamics are to depend on consciousness, consciousness needs to be dynamically relevant.%
    \footnote{This argument can be strengthened by considering what is required to distinguish two or more theories of consciousness empirically, cf.~\cite{kleiner2023closure}, where however dynamical relevance is referred to as `empirical version of the closure of the physical' in~\cite{kleiner2023closure}, and formulated in more generality than we do here.}

The upshot of these arguments is that dynamical relevance is a \emph{necessary condition} for the type of activities we carry out when engaging in \emph{empirical} scientific studies of consciousness. These arguments do not show that dynamical relevance is true. For all we know, there is the possibility that it isn't. But if it isn't, the empirical investigation of consciousness---and with it the science of consciousness---does not make sense; a necessary condition for its possibility would be violated.

\subsection*{Current Theories}
The above argument does not depend on any specific theory of consciousness. But it is interesting to ask what current theories of consciousness say about this point. What do they say about dynamical relevance?

First, it is important to note that empirical tests of theories of consciousness presume that consciousness is dynamically relevant according to these theories. That is the case, because they assume that whatever is measured can corroborate or falsify a theory, or speak in favour of one theory rather than another. For this to be possible, consciousness must make a difference to the data. And because the data is drawn from the physical dynamics of a system, consciousness must be dynamically relevant.

Second, we can consider the metaphysics of theories of consciousness. In the cases where these are clear, they do, in our eyes, imply dynamical relevance. Consider, as an example, Integrated Information Theory (IIT)~\cite{oizumi2014phenomenology}. IIT assumes that experience is primary and physics---or better, physical descriptions---are secondary. In a sense, only experience exists, in the form of cause-effect-structures. Hence it should be the case that experience makes a difference to the physical dynamics, so that conscious experience is dynamically relevant.  

Another example is Global Neuronal Workspace Theory (GNW)~\cite{dehaene2011global}. Here, too, we think, the metaphysical interpretation implies dynamical relevance. GNW assumes that conscious experiences are tied to a global neuronal workspace, ``consisting of a distributed set of (...) neurons characterised by their ability
to receive from and send back to homologous neurons in other (...) areas horizontal
projections through long-range excitatory axons''~\cite[p.\,56]{dehaene2011global}. Organisms that posses a workspace are conscious, while organisms that do not posses a workspace are not conscious, according to the theory. Hence whether or not a system is conscious makes a difference to a system's information processing architecture and, a fortiori, to the system's dynamics.

The only thing which speaks against dynamical relevance among current theories of consciousness, in our eyes, is their mathematical formulation (in those very limited cases where a mathematical formulation exists).

Consider, for example, Integrated Information Theory (IIT). The mathematics of IIT is given in terms of an unwieldy algorithm that takes as an input a physical description of a system \emph{as given by some physical theory}, and provides as output a mathematical description of the conscious experience of that system. An analysis of the mathematics that underlie this algorithm shows that the algorithm defines a map which goes from the physical descriptions to the descriptions of conscious experience~\cite{kleiner2021mathematical,tull2021integrated}.

Therefore, according to IIT's mathematics, consciousness is not dynamically relevant. The physical evolution of the systems are exactly as they are in the physical theory that provides the input to IIT. No change whatsoever is introduced to these dynamics by the theory. The mathematics of IIT do not instantiate dynamical relevance.

In our view, this is an issue of the mathematical formulation that IIT applies. The mathematics do not naturally align to the metaphysical foundation of the theory, and the exact same formal properties which speak against dynamical relevance are the source of other issues, most notably issues with falsifying the theory, cf.~\cite{kleiner2021falsification}, and issues related to the unfolding argument, more generally~\cite{doerig2019unfolding}. We think that the mathematics of IIT need to be revised, at the very least to instantiate dynamical relevance, so as to resolve the problems with falsification.

\section{Preliminaries}\label{sec:preliminaries}

The central notion which underlies our result is that of the time evolution of a system's states.  Given a scientific theory $T$ and a system $S$ within the scope of the theory, we denote by
$k_T(S,s)$ the \emph{dynamical evolution} (also called `trajectory') of $S$ with initial state $s$. This dynamical evolution describes how the state $s$ evolves in time according to $T$. An example is the evolution of a brain state according to a neuroscientific theory.
We will mostly abbreviate $k_T(S,s)$ by $k_T$ if it is clear from context that we're talking about one system and one initial state.

The class of theories which is relevant in the present context are physical theories, on the one hand, and theories of consciousness, on the other hand. We use the symbol $\Phys$ to denote physical theories that have been discovered by the natural sciences, in so far as they are relevant for AI or consciousness. Examples are theories of neuroscience, biology, chemistry, computer science and physics.

Different theories describe systems at different levels~\cite{list2019levels}, and in some cases, the states of a system posited by one theory $T$ (the ``lower'' level) can (in principle) be mapped to states of another theory $T'$ (the ``higher'' level). If this is the case, we write $T < T'$. Because dynamical evolutions are sequences of states, if $T < T'$, we can map any dynamical evolution $k_T$ of $T$ to a (not necessarily dynamical) evolution of $T'$, which we denote as $k_T|_{T'}$.

We assume that there is a physical theory $T_F \in \Phys$ that can be mapped to states of any other physical theory in $\Phys$, which means that $T_F < T $ for all $T \in \Phys$. For lack of a better term, we will refer to this theory as \emph{fundamental physical theory}, but emphasise, that it does not have to be ``the true'' fundamental theory. The requirement that $T_F < T $ for all $T \in \Phys$ is merely an epistemic requirement that expresses relationships been theories in $\Phys$, and leaves open whether $T_F$, or any other theory in $\Phys$ for that matter, is the true theory which correctly describes the actual dynamics. Whether or not this can be the case depends precisely on the question of whether consciousness is dynamically relevant. What justifies the assumption that there is a theory whose states can be mapped to states of the other theories (whose states \emph{ground} the states of all other theories, one might say) is that  the states of quantum theory can, in principle, be mapped to states of all physical theories in~$\Phys$. That is because quantum theory is what underlies condensed-matter theories as far as they are relevant for semi-conductors and integrated-circuit design of processors.
For all practical purposes, we can think of $T_F$ as quantum theory. We remark that the requirement of a relationship of states is much weaker than any reductive assumption.

Finally, we assume that there is a fact to the matter of what the real (that is: actual) dynamics of any system are, even if that fact may not be knowable. 
We denote the description of the real dynamics in terms of the states of any physical theory $T \in \Phys$ (any ``level'' of description, so to speak) by
$k^\ast|_T$. If $T<T'$, the description of the real dynamics in terms of the states of both theories are compatible, that is $k^\ast|_T|_{T'} = k^\ast|_{T'}$.

\section{Theories of Consciousness}\label{sec:tocs}

The second class of theories that are relevant in this context are theories of consciousness (tocs), which are sometimes also called models of consciousness. Tocs express a relation between a physical description of a system, on the one hand, and a description of its conscious experiences, on the other hand. The latter could be a description of its phenomenal character (cf. e.g.~\cite{kleiner2023mathematical,lee2021modeling}), or simply an expression of whether a system $S$ has conscious experiences at all. Together, the physical description and the description of conscious experiences applied by a toc $M$ constitute a state $s$ of the toc. Because a toc expresses a relation between a physical description of a system and a description of its conscious experiences, the state $s$ contains both a physical and an experiential part, which we refer to as \emph{physical} state and \emph{state of consciousness}, respectively; more about this below. The dynamical evolution $k_M(S,s)$ of a system $S$ in a state $s$ of the theory/model of consciousness $M$ expresses how the physical state and the state of consciousness relate according to the theory.

Because tocs contain a physical description of a system at some level, for every toc $M$, there is at least one physical theory $T_P \in \Phys$ such that the physical part of any state $s$ of $M$, and therefore also any dynamical evolution $k_M$, can be expressed in $T_P$. We denote this state by $s|_{T_P}$ and the expression of the physical part of the trajectory $k_M$ in terms of $T_P$ by $k_M|_{T_P}$. So, $k_M|_{T_P}$ is what $M$ says about the evolution of physical states on $T_P$'s level of description.%
    \footnote{
    We are grateful to an anonymous reviewer for pointing out that this might be a source of confusion. Here is another, more explicit, way of explain it. A theory of consciousness $M$ expresses a relation between a physical description of a system and a description of its conscious experience, that is a relation between a physical state and a state of consciousness. Let us suppose that the former constitute a set $\tilde P$ and that the latter constitute a set $E$. Here we are adding a $~$ on top of $P$ because the physical states which the theory of consciousness uses might not be identical to the physical states that any physical theory uses; there could be simplifications, for example. What needs to be the case, however, is that these states can be mapped to the states of \emph{some} physical theory $T_P$. The states of the physical theory are what the theory of consciousness ``means'' when addressing physical states, so to speak. Let us assume that the physical states of $T_P$ form a set. A trajectory $k_M$ of $M$ is a trajectory over $\tilde P \times E$, in this notation. By restricting to $\tilde P$ and then mapping to $P$, we obtain a trajectory over $P$. This is what the symbol $k_M|_{T_P}$ denotes: it is what the trajectory of $M$ implies for the physical time evolution when expressed in terms of the states of the theory $T_P$. 
    }
We call any such $T_P$ an \emph{underlying} physical theory of~$M$.

Independently of what the description is that a toc applies on the side of consciousness, there is a fact to the matter of whether a system is conscious or not when in $k_M(S,s)$. This means: whether the system $S$ has conscious experiences at least at one point of time in the dynamical evolution~$k_M(S,s)$.
Making use of the important link between tocs and physical descriptions, we can say that a system $S$
\emph{is conscious} in a physical evolution $k_{T_P}$ iff there is a dynamical evolution $k_M$ of $M$ such that (a) we have $k_M|_{T_P} = k_{T_P}$ and (b) the system is conscious in $k_M$.

Whether a toc has anything original to say about the dynamical evolution of its physical states, or simply presumes the dynamical evolution of an underlying physical theory, is precisely the question of dynamical relevance, defined as follows. Let $M$ denote a toc and $T_P \in \Phys$ an underlying physical theory of~$M$.

\begin{Def}\em\label{def:CDRtheory}
\emph{Consciousness is dynamically relevant} according to $M$ with respect to $T_P$ iff
\begin{equation*}
S \textrm{ is conscious in } k_M \ \Rightarrow \ k_M|_{T_P} \neq k_{T_P} \:.
\end{equation*}
\end{Def}
Here, the right-hand-side is short-hand for $k_M(S,s)|_{T_P} \neq k_{T_P}(S,s|_{T_P})$, where 
$s|_{T_P}$ denotes the restriction of the state $s$ of $M$ to $T_P$. The left-hand side is a shorthand for `$S$ is conscious in $k_M(S,s)$', meaning that there is at least one point of time in $k_M(S,s)$ so that $S$ has a conscious experience at that time according to $M$. The definition expresses the intuition that if $S$ is conscious according to a toc~$M$, then the dynamical evolution as specified by~$M$ differs from the dynamical evolution as specified by the underlying physical theory alone.

We have already referenced the `real' dynamics of a system and introduced the symbol $k^\ast|_{T_P}$ to denote what the real dynamics of a system would look like in terms of the states of $T_P$. There is also a fact to the matter of whether a system in a trajectory $k^\ast$ is conscious and how conscious experiences relate to the physical. That is, there is a `true' or `real' theory of consciousness, which we denote by $M^\ast$. As in the physical case, $M^\ast$ may be unknown and or unknowable. We will denote its dynamical evolutions by $k_{M^\ast}$.
Because these describe what really happens, we have $k_{M^\ast}|_{T_P} = k^\ast|_{T_P}$ for all~$T_P$.
Using $M^\ast$, we can define dynamical relevance simpliciter:

\begin{Def}\em\label{def:CDR}
\emph{Consciousness is dynamically relevant} (CDR) only if it is dynamically relevant according to the `true' toc $M^\ast$
with respect to \emph{some} physical theory $T_P \in \Phys$.
\end{Def}

\section{Verification}\label{sec:AI-sys}

What is unique about AI systems in the present context is not the particular architecture that is employed; AI can also be built on architecture derived from the brain; cf. e.g ~\cite{friston2022designing}. What is unique is rather that the architecture runs on CPUs, GPUs, TPUs or other processors that have been designed and \emph{verified} in the lab.

There are two major verification steps in processor development, called functional and post-silicon verification. \emph{Functional verification}~\cite{mishra2005functional,wile2005comprehensive} is applied once the design of a processor in terms of integrated circuits has been laid out, but before the manufacturing phase begins.
It applies simulation tools, formal verification tools and hardware emulation tools to ensure that the design of the chip meets the intended specifications as described by a computational theory $\Tcomp$. 
 \emph{Post-silicon verification}~\cite{mishra2017post,mitra2010post} is applied after the silicon waver has been fabricated. It applies in-circuit testing, functional testers, failure analysis tools and reliability testing, among other things, to ensure that the physical product works as $\Tcomp$ would have it.

Functional verification is a theoretical endeavour. It applies simulation and emulation tools based on a theoretical account on how the substrate, on which a processor is to be built, behaves. Because this substrate is a semi-conductor, this theoretical account is based on quantum theory $T_F$.
Put in terms of dynamics, functional verification aims to ensure that 
whatever happens in the quantum realm implements or is compatible with the dynamics as described by $\Tcomp$, formally:
\begin{equation}\label{eq:func_veri}
k_{T_F}|_{\Tcomp} = k_\Tcomp
\end{equation}
for all dynamical evolutions of a processor~$S$.

Post-silicon verification, on the other hand, is applied to a chip once it has been built. It ensures that the dynamics of the actual physical product comply with $\Tcomp$. Making use of the $k^\ast$ notation to denote the actual dynamical evolution of a system, post-silicon verification enforces that
\begin{equation}\label{eq:post_veri}
k^\ast|_{\Tcomp} = k_\Tcomp
\end{equation}
for all dynamical evolutions of a processor~$S$.

Being an AI system means running on CPUs, GPUs, TPUs or other processors
that have been designed and verified. That's what makes the system ``artificial''. 
And because processor dynamics compose (the output of one is the input of the next), verification holds for AI systems as well: there is an underlying computational theory $\Tcomp$ that accounts for what ``happens'' on the processors while the system is running, and the computational dynamics satisfy~\eqref{eq:func_veri} and~\eqref{eq:post_veri}. 

\section{AI Consciousness}\label{sec:AIconsc}

With all this in place, we can formulate the question that is being asked precisely.
The term `artificial intelligence' is used very broadly, comprising many different computational architectures and applications. What one means when one asks whether an AI system is conscious is whether the computational architecture that is applied by this system, with the specific quirks of its implementation and training, potentially in a specific task, has conscious experiences. The architecture and these specifics determine the computational dynamics the system is capable of. Thus, the question is whether the system has a computational evolution $k_\Tcomp$ such that it is conscious in this computational evolution according to a theory of consciousness $M$; cf. Section~\ref{sec:tocs} for a definition of what this means in terms of dynamics $k_M$ of $M$.%
    \footnote{The point here is to restrict downwards, not upwards. Any question ``above'' the computational level can be posed in terms of computational dynamics.
    }
In summary:

\begin{Def}\em\label{def:AI-sys-conscious}
An \emph{AI system $S$ is conscious} according to a theory of consciousness $M$ only if 
there is at least one dynamical evolution $k_\Tcomp$ in which the system is conscious according to $M$.
\end{Def}

This is a very weak condition, which however has one important consequence: that the question of AI consciousness is determined by facts on the computational level and above; it is independent of what happens on a sub-computational level. 
That is, if we have a a trajectory $k_{T_P}$ on a sub-computational level ($T_P < \Tcomp$) 
with $k_{T_P}|_{\Tcomp} = k_{\Tcomp}$
then $S$ is conscious in $k_\Tcomp$ only if
it is conscious in $k_{T_P}$.

\section{Main Result}

Our main result is the following theorem.

\begin{Thm}\em\label{thm}
If consciousness is dynamically relevant, then AI systems aren't conscious.
\end{Thm}

Before giving the proof, we first illustrate the result for the 
simpler case where consciousness is dynamically relevant with respect to the computational level $\Tcomp$ itself. The power of the theorem is to extend this result to all other cases. Subsequent to this illustration, we prove a lemma needed for the main theorem, and then proceed to prove the theorem itself.

So let us consider the case where $T_P$ in Definition~\ref{def:CDR} is $\Tcomp$. 
Let $S$ be an AI system. Because of post-silicon verification~\eqref{eq:post_veri}, all of the dynamical
evolutions of $S$ satisfy 
\begin{equation}\label{eq:proof3}
    k^\ast|_{\Tcomp} = k_\Tcomp \:.
\end{equation}
Application of Definition~\ref{def:CDR} for the case $T_P = \Tcomp$ implies,
via~Definition~\ref{def:CDRtheory}, that if $S$ is conscious in a $k_{M^\ast}$, then $k_{M^\ast}|_{\Tcomp} \neq k_{\Tcomp}$. The converse of this statement is that if $k_{M^\ast}|_{\Tcomp} = k_{\Tcomp}$,
then $S$ is not conscious in  $k_{M^\ast}$.
Because $k_{M^\ast}|_{\Tcomp} =  k^\ast|_{\Tcomp}$, the identity \eqref{eq:proof3} establishes the prerequisite of this condition for all dynamical evolutions of $S$. Therefore, it follows that $S$ is not conscious in any $k_{M^\ast}$. Thus, Definition~\ref{def:AI-sys-conscious} implies that $S$ is not conscious, as claimed.

The remainder of this section is devoted to the proof of the theorem in the general case. To this end, we first state and prove the following lemma. 

\begin{Lemma}\em\label{lem:passes-downwards}
\emph{Dynamical relevance passes downward}, in the sense that if $T_P < T_P'$ and consciousness is dynamically relevant according to $M$ with respect to $T_P'$, then it is also dynamically relevant according to $M$ with respect to $T_P$.
\end{Lemma}

\Proof[Proof of the Lemma]
Consciousness is dynamically relevant according to $M$ with respect to $T_P'$, iff
\begin{equation*}
S \textrm{ is conscious in } k_M  \ \Rightarrow \ k_M|_{T_P'} \neq k_{T_P'} \:.
\end{equation*}
Because $T_P < T_P'$, there is a function which maps states---and therefore also dynamical evolutions---from $T_P$ onto $T_P'$.
Therefore, we have
\begin{equation*}
k_M|_{T_P'} \neq k_{T_P'} \ \Rightarrow \ k_M|_{T_P} \neq k_{T_P} \:.
\end{equation*}
Together with the above, this gives
\begin{equation*}
S \textrm{ is conscious in } k_M \ \Rightarrow \ k_M|_{T_P} \neq k_{T_P} \:,
\end{equation*}
which is the case iff consciousness is dynamically relevant according to $M$ with respect to $T_P$.
\QED

We now proceed to the proof of the theorem.

\Proof[Proof of the Theorem]
We first consider the case where $T_P$ in Definition~\ref{def:CDR} is $T_F$.

Let $S$ be an AI system.
Because of functional and post-silicon verification, we have
\begin{equation}\label{eq:proof1}
k_{T_F}|_{\Tcomp} = k_\Tcomp = k^\ast|_{\Tcomp}
\end{equation}
for all dynamical evolutions of $S$.
Because consciousness is (by assumption) dynamically relevant and we have assumed $T_P = T_F$, Definition~\ref{def:CDRtheory} applies to give
\begin{equation}\label{eq:proof2}
S \textrm{ is conscious in } k_{M^\ast} \ \Rightarrow \ k_{M^\ast}|_{T_F} \neq k_{T_F}
\end{equation}
for all dynamical trajectories $k_{M^\ast}$ of $M^\ast$.

Let us now assume that $S$ is conscious in some trajectory $k_{M^\ast}$ of $M^\ast$.
According to the last implication, we thus have
\begin{equation*}
     k_{M^\ast}|_{T_F} \neq k_{T_F} \:.
\end{equation*}
Because $T_F < \Tcomp$, we can map both of these
trajectories to $\Tcomp$. For $k_{M^\ast}|_{T_F}$, this gives
\begin{align*}
    k_{M^\ast}|_{T_F}|_{\Tcomp} &= k^\ast |_{T_F}|_{\Tcomp} \\
    &= k^\ast|_{\Tcomp} = k_{M^\ast} |_{\Tcomp} \:,
\end{align*}
where we have made use of identities established
in Sections~\ref{sec:preliminaries} and~\ref{sec:tocs}.
Equation~\eqref{eq:proof1} furthermore establishes that
\begin{equation*}
     k_{M^\ast} |_{\Tcomp} = k^\ast|_{\Tcomp} = k_\Tcomp \:.
\end{equation*}
The two facts that (a) $k_{M^\ast}|_{\Tcomp} = k_\Tcomp$ and
(b) that $S$ is conscious in $k_{M^\ast}$ establish that
$S$ is conscious in $k_\Tcomp$.

Equation~\eqref{eq:proof1} also establishes that
\begin{equation*}
    k_{T_F}|_{\Tcomp} = k_\Tcomp \:.
\end{equation*}
Because of this equation and $T_F < \Tcomp$, the implication
of Definition~\ref{def:AI-sys-conscious} explained in the last
paragraph of Section~\ref{sec:AIconsc} applies and establishes
that $S$ is conscious in
$k_{T_F}$.

Unwrapping what `$S$ is conscious in
$k_{T_F}$' means by definition, we find that
there must be a dynamical evolution
$\tilde k_{M^\ast}$ of $M^\ast$ such that
\begin{align*}
    &\textrm{(a) } \tilde k_{M^\ast}|_{T_F} = k_{T_F} \quad \textrm{and}\\
    &\textrm{(b) } S \textrm{ is conscious in } \tilde k_{M^\ast} \:.
\end{align*}
Together, these two conditions violate~\eqref{eq:proof2}.
Thus we have arrived at a contradiction.

The assumptions that went into the derivation of this contradiction were that
consciousness is dynamically relevant with respect to the $T_F$ level, that $S$
is an AI system, and that $S$ is conscious in a trajectory $k_{M^\ast}$ of $M$.
The first assumption is stated as a condition in the theorem. Thus it follows that
the latter two cannot be both the case.

Because $k_{M^\ast}$ was arbitrary,
it follows that an AI system $S$ cannot be conscious in any
trajectory $k_{M^\ast}$ of $M^\ast$. Consequently, applying Definition~\ref{def:AI-sys-conscious},
it cannot be conscious at all.
This establishes the claim that if consciousness is
dynamically relevant with respect to $T_F$, then AI systems aren't conscious.

It remains to consider all other cases of $T_P$ in Definition~\ref{def:CDR}.
Therefore, let us assume that consciousness is dynamically relevant with respect to some $T_P \neq T_F$.
Because $T_F < T_P$ for all $T_P \in \Phys$, and because dynamical relevance passes downward
(Lemma~\ref{lem:passes-downwards}), it follows that consciousness is also dynamically relevant with respect
to $T_F$. Hence the previous case applies and the result follows in full generality.
\QED

\section{Objections} 

In this section, we discuss a few immediate responses to our result.

\subsection{Verification is imperfect}
Verification is an industrial process that may not be perfect: despite functional and post-silicon verification, the actual dynamics of a processor may not adhere to the computational theory targeted by verification in all cases. Verification may leave a bit of wiggle-room for the dynamics to diverge from the computational theory. Could this wiggle-room suffice for consciousness to unfold its dynamical effects?

Any answer to this question depends on how exactly consciousness is dynamically relevant and which imperfections arise in day-to-day verification. It is natural to expect that consciousness' dynamical relevance is systematic in nature: dynamical effects should systematically occur if a system is conscious and make a systematic difference to how the system evolves in time. The imperfections in day-to-day verification, on the other hand, are likely to be mostly random in nature, meaning that the deviation in dynamical evolution they fail to suppress are random too, both in time (when such a deviation can occur) and in the extend to which they can make a difference. If this is true, it is unlikely that the wiggle-room left open due to imperfections suffices for consciousness to unfold its dynamical effects.

\subsection{Determinism}
One objection to our result takes our result to show or imply that a deterministic system cannot be conscious, and argues that this is very unlike to be true. Hence the result must be wrong or rest on very weak assumptions, so the objection goes.

This objection fails because our result does not show or imply that deterministic systems cannot be conscious. 
What prevents a system from being conscious, according to our result, is that its design forces it to comply to a formal system that is independent of consciousness. The system is ``locked into'' a formal system, so to speak. It cannot deviate from it. Reality is forced to adhere to a theoretical construct, by design.

Our result is fully compatible with deterministic systems, and also with a deterministic relevance of consciousness to a system's dynamics.

\subsection{Probabilistic processing}\label{sec:prob-proc}
Verification as applied in industry targets deterministic computational theories. Would our result also hold in case of verified probabilistic processing?

The mathematical framework we apply is compatible with probabilistic processing: we do not make an assumption as to whether the notions of state and dynamical evolution are deterministic or not; a state may well be a probability distribution and its dynamical evolution a stochastic process.
Verification, in this case, implies that a system conforms to the stochastic process as described by a stochastic computational theory. This leaves room for consciousness to have a dynamical effect, but only if this effect conforms to the probability distributions as described by the stochastic computational theory.
That is, consciousness may determine how the probability distributions of the stochastic computational theory are sampled, 
but it cannot change them. As in the case of imperfect verification, we remain sceptical as to whether this limited freedom is compatible with the systematic nature of consciousness' dynamical effects that are to be expected.

\subsection{Quantum computing}
Does our result also hold true in the case of quantum computing? Quantum computing is a young industry and it is not yet clear which type of verification, if any, will need to be deployed. It is likely, however, that any type of verification will need to presuppose a notion of \emph{measurement}, which is an inherently vague concept in quantum theory~\cite{bell1990against} that is partially external to the account of quantum dynamics by the Schrödinger equation. If consciousness were related to measurement (for example via consciousness-induced dynamical collapse as proposed in~\cite{chalmers2021consciousness}), then verification might leave enough room for consciousness to
have a systematic and meaningful effect.
If, on the other hand, consciousness is not related to measurement in quantum theory, it is likely that verification of quantum computers to adhere to quantum dynamics will preclude any potential dynamical effects of consciousness; just as in the classical case.

\section{Conclusion}
This paper addresses the question of whether AI systems are conscious. 
Its objective is to introduce a new formal tool, in the form of a theorem,
that provides an answer to this question which is independent of the specific computational
architecture that an AI system utilises, and which does not rely on any specific cognitive feature that an AI
system might possess or lack that may be related to conscious experience.


Our result is based on what we take to be the only property that distinguishes AI systems from other cognitive systems, a property that might well embody the actual meaning of the word `artificial' in `artificial intelligence': that the system runs on a substrate that has been designed and verified, rather than naturally evolved. 

Ultimately, we believe that any scientific statement about whether a system is conscious needs to be based on a theory of consciousness that is supported by theoretical, philosophical, and most importantly empirical evidence. The Science of Consciousness%
    \footnote{Also called \emph{Scientific Study of Consciousness} to emphasise the importance of contributions from humanities, most notably philosophy.}
searches for such theories. The crucial premise in our result---dynamical relevance---is a \emph{property} which theories ascribe to consciousness, so that our theorem can be regarded as establishing a fact about AI's capability for consciousness for a whole class of theories of consciousness: all those that posit consciousness to be dynamically relevant.
Results of this form are important as long as evidence in favour of any single theory of consciousness, as well as evidence to distinguish among them, is still in its early stages, and while the space of possible theories remains only partially explored.

Our result has a few interesting, slightly funny, and potentially relevant implications for AI engineering and AI interpretability. The most notable of these is that our result shows that \emph{if an AI system states that it is conscious, then this cannot be because it is conscious.} That is to say, even if an AI system were conscious, the cause of any such statement cannot be that the AI system is conscious. This follows because if such a cause existed, consciousness would have to be dynamically relevant, in which case our theorem implies that the system isn't conscious.
Another implication is that if consciousness has functions that could improve a system's information processing, then, to make use of those functions, theories of consciousness should be taken into account when designing the substrate on which an AI system will run. 

The question of whether AI systems are conscious is of major societal concern~\cite{amcs2023openletter}.
It has important ethical~\cite{bostrom2018ethics,metzinger2021artificial}, legal~\cite{benzmuller2020reasonable,susskind2019online}, and technological consequences,
and will likely play a major role in shaping governance of AI and how individuals interact with this technology. Our result aims to deliver a rigorous and justified answer to this question that does not rely on contingent assumptions, such as the truth of a particular theory of consciousness, or the validity of a particular test of consciousness when applied to AI systems.

\subsection*{Acknowledgements}
We would like to thank the participants of the \emph{Modelling Consciousness Workshops 2022 and 2023} of the \emph{Association for Mathematical Consciousness Science} for valuable discussions on the topic of AI consciousness and feedback about this result, specifically Alexandra Proca, Cameron Beebe, Sophie Taylor, Peter Thestrup Waade,  Joscha Bach, Mathias Gutmann, George Deane, Jordan O'Byrne, Ian Durham and Sean Tull, Wanja Wiese, Stephan Sellmaier, Jonathan Mason and Justin Sampson for feedback on an earlier version of this manuscript.
This research was supported by grant number FQXi-RFP-CPW-2018 from the Foundational Questions Institute and Fetzer Franklin Fund, a donor advised fund of the Silicon Valley Community Foundation. We would like to thank the Dutch Research Council (NWO) for (partly) financing TL's work on project number 182.069 of the research programme Fluid Spintronics, and the Mathematical Institute of the University of Oxford for hosting JK while working on this project.

\subsection*{Statement of Data Availability}
No new data were generated or analysed in support of this research.

\end{multicols}
\end{document}